\documentclass[10pt,twocolumn,letterpaper]{article}

\usepackage[pagenumbers]{cvpr}

\usepackage{cuted}
\usepackage{currfile}
\usepackage{caption}
\usepackage{booktabs}
\usepackage{tabularx}
\usepackage{multirow}
\usepackage{makecell}
\usepackage{xspace}
\usepackage{microtype}
\usepackage{needspace}
\newcolumntype{Y}{>{\raggedright\arraybackslash}X}
\newcolumntype{L}[1]{>{\raggedright\arraybackslash}p{#1}}

\newcommand{\takeaway}[1]{\par\smallskip\noindent\textbf{Takeaway.} \emph{#1}\par}

\newcommand{\ci}[2]{\,[\ensuremath{#1},\,\ensuremath{#2}]}
\newcommand{\gem}{Gemini\xspace}

\newcommand{\opus}{Opus\xspace}
\newcommand{\sol}{Sol\xspace}

\makeatletter
\renewcommand\paragraph{\@startsection{paragraph}{4}{\z@}{1.2ex \@plus .3ex \@minus .2ex}{-1em}{\normalfont\normalsize\bfseries}}
\makeatother
\usepackage{placeins}

\definecolor{cvprblue}{rgb}{0.21,0.49,0.74}
\usepackage[pagebackref,breaklinks,colorlinks,allcolors=cvprblue]{hyperref}

\def\paperID{}
\def\confName{}
\def\confYear{}

\title{Telling Is Not Seeing: Textual Personas Rarely Make Vision--Language Models Answer Like the People They Describe}

\author{Jiawei Chen\\
Netflix\\
{\tt\small jiaweic@netflix.com}}

\begin{document}
\maketitle

\begin{abstract}
Vision--language models (VLMs) are increasingly prompted with personas, such as ``You have low vision'', to stand in for people in user studies. We ask whether a persona makes a model see like the person it describes. We compare persona answers with measured people with low vision, colour blindness or hearing loss, and with simple rules that use only the facts the persona states. The persona never beats these rules, and it erases the real differences between groups and individuals. Describing blur leaves recognition unchanged in all 14 models we test, while blurring the photo lowers it as it does for people. Inside open models, the description changes neither the image information nor the answer probabilities. Yet the models know the effects they fail to show: they can estimate how blur and illusions affect people. Telling is not seeing: a persona makes a model say what it knows about people, not see what they see.
\end{abstract}

\begin{figure}[!h]
\centering
\includegraphics[width=\columnwidth]{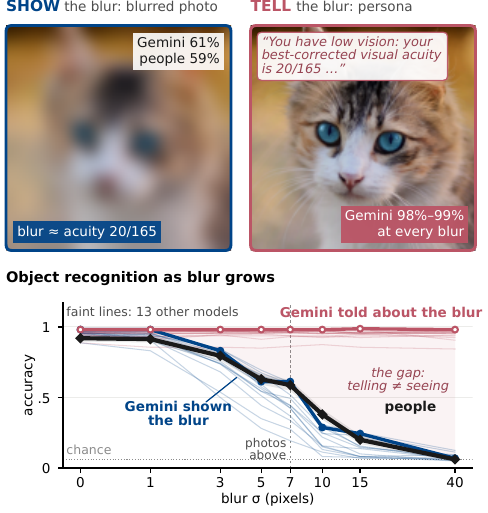}
\vspace{-1.9em}
\caption{\textbf{Telling a model about blur does not make it see blur.} Four sighted people and \gem name objects in photos at eight blur levels. Top: \gem is either shown the blurred photo, or given the clean photo and told about the blur by a persona that states the matching acuity. Bottom: shown the blur, \gem loses accuracy as people do; told about it, \gem stays at 98--99\%. The photos illustrate the blur levels; the test photos are greyscale.}
\label{fig:teaser}
\vspace{-1.3em}
\end{figure}

\section{Introduction}\label{sec:intro}
Designers need to know whether a person with low vision can read a sign, or a deaf viewer can follow fast subtitles, and user studies are slow and costly. A vision--language model (VLM) with a persona prompt, such as ``You have low vision,'' is an attractive replacement. Language models already stand in for survey respondents and users~\cite{argyle2023,aher2023,park2024,matraix2026}. This paper asks whether a persona makes a VLM answer like the people it describes.

Earlier work gives mixed answers. In a study of 40 people with low vision, vision information in the prompt did not bring GPT-4o closer to their answers~\cite{natalie2025}. A textual low-vision persona did not lower a VLM's reading performance to the expected level~\cite{gao2025viocr}. Colour-blind personas still read Ishihara plates (dot patterns that hide a digit) like a person with normal colour vision~\cite{hayashi2026cvd}. These studies compare a persona with a minimal prompt (no persona) or with an answer key. When a persona helps, this cannot show whether the model simulates the person or the persona simply contains a useful fact, such as a stated acuity. None of these studies tests whether personas keep the measured differences between groups.

We test both points. \emph{The key distinction is between knowing facts about a viewer and answering as that viewer.} For each persona, we also make the simplest prediction from its stated facts, with no persona prompt (a \emph{task-specific control}): a clinical rule or a textbook answer that uses only the stated facts, or the model without the persona. A persona that simulates a person must at least match it. We also compare personas with several measured groups, and test whether the personas keep the differences between these groups (\cref{tab:main}).

Current frontier models do change their reading and colour answers with a persona (\cref{fig:reading,fig:d15}), but not toward the people. A blur test suggests why (\cref{fig:teaser}), and illusion tests show that models know effects that their own answers lack.

\paragraph{Contributions.}
\begin{enumerate}[nosep,leftmargin=*]
\item \textbf{An evaluation of personas against measured people.} We compare personas with measured people: readers with low vision, colour-blind viewers, and subtitle viewers in five groups, including deaf and hard-of-hearing viewers. Each persona must at least match a task-specific control that uses only the facts the persona states, and must keep the differences between groups and individuals. No persona passes these tests.
\item \textbf{A controlled test of telling versus showing.} On the same photos, we compare a description of blur with the blurred photo itself, against human effect sizes, in 14 VLMs. The description has no effect in any model; the blurred photo gives seven models a loss as large as people's.
\item \textbf{An inspection of the models' internals.} Through answer probabilities and linear probes of the hidden states of open models, we show that the persona changes neither the image information nor the decision, and adds no blur signal.
\item \textbf{A separation of what models know from what they show.} On the same images, the models' estimates of people's answers recover most of the human illusion and blur effects, while their own answers show little or none of them.
\end{enumerate}

\begin{table}[t]
\centering\scriptsize
\setlength{\tabcolsep}{2pt}\renewcommand{\arraystretch}{1.1}
\caption{\textbf{Overview of the five studies.} Frontier models: Gemini-3.8-Flash, Claude-Opus-5.5, GPT-6-Sol. $^a$Three of the four open models cannot do the task without a persona. $^b$One open model cannot run the subtitle-speed clips.}
\label{tab:main}
\vspace{-0.6em}
\begin{tabularx}{\columnwidth}{@{}L{1.55cm}L{1.35cm}L{1.55cm}YL{0.95cm}@{}}
\toprule
Study: task & People (data) & The persona states & We compare with & Models \\
\midrule
\multicolumn{5}{@{}l}{\emph{Primary tests: people whose own answers are recorded} (\cref{sec:measured})} \\
Low-vision reading: read sentences whose print size is stated & 56 readers with low vision~\cite{xiong2021lowvision} & The reader's letter-chart acuity, contrast sensitivity and diagnosis & The rule ``reading limit = letter-chart acuity''; an image filter & 3 frontier, 4 open \\
\cmidrule(lr){1-5}
Colour sorting (D-15): order 15 caps by hue & 39 colour-blind viewers (group scores)~\cite{vingrys1988d15} & The type of colour blindness & No persona; the textbook order for the stated type; colours changed for that type & 3 frontier, 4 open$^a$ \\
\cmidrule(lr){1-5}
Subtitle viewers: choose line breaks; rate clips; state preferences & 88--94 viewers in five groups~\cite{szarkowska2018suredata} & The participant's language, age, gender, English level and hearing & The same model without a persona & 3 frontier, 4 open$^b$ \\
\midrule
\multicolumn{5}{@{}l}{\emph{Controlled test: sighted people view changed images} (\cref{sec:proxy})} \\
Blur: name the object in a photo & 4 observers with normal vision~\cite{geirhos2021} & The acuity that matches the blur & The blurred photo, no persona & 14 \\
\midrule
\multicolumn{5}{@{}l}{\emph{Diagnostic test: the model estimates a group's answers} (\cref{sec:stated})} \\
Illusions: choose the longer line or larger disc & 256 people~\cite{makowski2023illusion} & An age group (optional) & The model's estimate of the share of people who choose each option & 3 frontier; 11 open (4 types) \\
\bottomrule
\end{tabularx}
\vspace{-1.0em}
\end{table}

\section{Related Work}\label{sec:related}
\paragraph{Personas and synthetic participants.} Persona prompts can caricature groups~\cite{cheng2023compost}, flatten the differences within them~\cite{wang2025nmi} and misrepresent their opinions~\cite{santurkar2023}. Language models describe a group's opinions better than they answer as its members~\cite{meister2025distributional}. Personas can also change what a model says about itself more than what it does~\cite{han2025}, and can trigger refusals or biased reasoning~\cite{gupta2024bias}, including on vision tasks~\cite{salewski2023}. Synthetic participants have been tested on images, video and visual marketing~\cite{tsai2026seeing,shrestha2026}. Simulated-user platforms select cohorts by accessibility needs and check that persona agents follow their stated attributes~\cite{matraix2026}. Following an attribute is not answering like the person: our reading personas follow the stated acuity almost exactly, yet cannot beat the acuity rule. Natalie \etal~\cite{natalie2025} built one persona for each of 40 participants with low vision: vision information alone left GPT-4o's agreement with them at 0.59, as with a minimal prompt, and adding one example answer raised it to 0.70. We add measured people, changed colours, task-specific controls, and the differences between groups.

\paragraph{VLMs compared with human vision.} VLMs struggle with simple geometry~\cite{rahmanzadehgervi2024blind}, basic visual patterns~\cite{tong2024eyes}, core perception tasks~\cite{fu2024blink}, human-like visual search~\cite{lin2024hvsbench}, colour~\cite{liang2025colorbench} and contrast sensitivity~\cite{hernandez2025csf}. Human data on distorted images~\cite{geirhos2018,geirhos2021,geirhos2020kappa,geirhos2019shape,huber2023children} show where models differ from people. A filter that copies human contrast sensitivity makes the errors of image classifiers more human-like~\cite{wolff2026lowpass}. Illusion benchmarks compare model and human answers~\cite{zhang2023illusions,shahgir2024illusionvqa}, and models often answer from memory of a famous illusion~\cite{guan2024hallusionbench,ullman2024illusion}. We compare a model's own answers with its estimates of how people answered on the same images.

\section{Method}\label{sec:protocol}
\paragraph{Three kinds of answer.} For each item, we use the question that the people answered. The three kinds of answer differ in who answers the question and in what the model receives.
\begin{itemize}[nosep,leftmargin=*]
\item A \emph{persona answer}: the model answers the question itself, as the person that the persona describes. It sees the original image or video and the persona. We score its answer as we score a participant's answer. For example, it sees a clean photo with the persona ``your best-corrected visual acuity is 20/165'' and names the object as that person would.
\item A \emph{changed-image answer}: the model answers the question itself, with no persona, but it sees a changed image. For example, it sees the photo blurred to match that acuity.
\item An \emph{estimate}: the model does not answer the question. It stays itself and estimates, as a number, how a group of people answered it, not how one person answered. For example, it sees an illusion image and reports the percentage of adults who chose the left line.
\end{itemize}
Thus, a persona answer tests whether the model can answer as a person, and an estimate tests whether the model knows how a group of people answers. A persona that makes a model answer like a person should change the persona answer in the same way as the person's answer changes. An estimate can be correct even if the model does not see the image as people do. Thus, we keep estimates apart from the model's own answers.
Formally, let $i$ index an item (an image, a sentence or a video), $g$ a viewer or viewing condition with $g=0$ the reference, $M$ the model, $x_i$ the original input, $p_0$ a neutral prompt and $p_g$ the persona of $g$. The three kinds of answer are
\begin{equation}\label{eq:answers}
\begin{gathered}
A_i^{P}(g)=M(x_i,p_g),\qquad A_i^{T}(g)=M\big(T_g(x_i),p_0\big),\\
\hat{Y}_i(g)=M(x_i,q_g),
\end{gathered}
\end{equation}
where $T_g$ changes the input as condition $g$ would (for example, a blur), and $q_g$ asks the model to estimate how people in group $g$ answered item $i$. The answer without a persona is $A_i^{0}=M(x_i,p_0)$. Only one part of the input differs between $A_i^0$ and $A_i^P(g)$: the text.

\paragraph{The persona must match a task-specific control.} A persona can help for a simple reason: it states a useful fact. A reader's letter-chart acuity, for example, is a good guess of the smallest print that the reader can read. So, for each study, we compare the persona with the simplest use of its facts. In reading, we copy the stated acuity as the predicted reading limit; we call this control the \emph{acuity rule}, a clinician's first guess. The persona also states the contrast sensitivity and the diagnosis, and the model knows that macular disease makes reading harder than letter acuity suggests, so a model that simulates the reader should move away from the rule in the right direction. In colour sorting, the controls are the textbook order for the stated type and the model without a persona; in the subtitle tasks, the control is the model without a persona. Each control uses no more than the persona states.

\paragraph{How we compare model and human effects.} Each test has a reference condition, such as a clean photo, and a test condition, such as a blurred photo. The \emph{human effect} is the change in people's average score from the reference to the test condition. The \emph{model effect} is the same change in the model's score. The \emph{relative effect} is the model effect divided by the human effect. A value of 1 means that the model changes as much as people; 0 means that it does not change. For example, if blur lowers people's accuracy by 50 points and a model's accuracy by 25 points, the relative effect is 0.5. Formally, let $s_i(\cdot)$ score an answer (for example, correct or not, or a rating) and let $Y_i^{H}(g)$ be the human answers. Then
\begin{gather}
\Delta_H(g)=\mathbb{E}_i\big[s_i(Y_i^{H}(g))-s_i(Y_i^{H}(0))\big],\label{eq:rel}\\
R(g)=\Delta_M(g)/\Delta_H(g),\notag
\end{gather}
where $\Delta_M(g)$ is defined in the same way for a kind of model answer in \cref{eq:answers}. When a test has several levels, such as blur strengths, we sum both effects over the levels. We call a model effect \emph{as large as people's} when the 90\% confidence interval of $R$ lies in $[0.80,1.25]$ (an equivalence test~\cite{lakens2017tost}; a 90\% interval gives a 5\% error rate for this test). We say that a persona has \emph{no effect} when this interval lies in $[-0.20,0.20]$. Because we test many models, we use the Holm correction~\cite{holm1979}, which keeps the chance of any false claim across the models at 5\%. Square brackets give 95\% confidence intervals, computed by resampling items and people (bootstrap).

\paragraph{Same average, different errors.} A model and a person can both miss 40 of 100 photos but miss different photos. Thus, we also report error agreement ($\kappa$~\cite{geirhos2020kappa}): 0 means that the two miss photos independently, and 1 means that they miss the same photos. Formally, if $c_{\mathrm{obs}}$ is the share of photos on which model and person are both right or both wrong, and $a_M$ and $a_H$ are their accuracies, then
\begin{equation}\label{eq:kappa}
\kappa=\frac{c_{\mathrm{obs}}-c_{\mathrm{exp}}}{1-c_{\mathrm{exp}}},\qquad c_{\mathrm{exp}}=a_M a_H+(1-a_M)(1-a_H).
\end{equation}

\paragraph{Models.} The core panel is three frontier VLMs, proprietary models that we call through their providers' interfaces at their default temperature: Gemini-3.8-Flash (\gem), Claude-Opus-5.5 (\opus) and GPT-6-Sol (\sol). We also run open-weight models from seven families~\cite{bai2025qwen25vl,bai2025qwen3vl,xu2025qwen3omni,wang2025internvl35,zhu2025internvl3,li2024llavaov,clark2026molmo2,gemma2026gemma4,phi4mini2025,nvidia2026nemotronomni}, without fine-tuning. One rule applies to every study. The core panel and the same four open models (Qwen3-VL-8B, Qwen2.5-VL-72B, InternVL3.5-8B, Gemma-4-12B) run every primary image task. A persona result counts only if the model can do the task without a persona. Video tasks use only models that accept the input; \opus and \sol take no video, so they get time-labelled frames. The blur test also runs seven more open models (14 in total). Each persona answer is one sample. Repeating 10\% of the calls shows that this is stable: a persona's reading limit changes by less than one chart line (0.03 to 0.07\,logMAR), and the personas still barely differ from each other in the subtitle task (supplement).

\section{Results}\label{sec:results}

\subsection{Do personas answer like measured people?}\label{sec:measured}

\begin{figure}[t]
\centering
\includegraphics[width=\columnwidth]{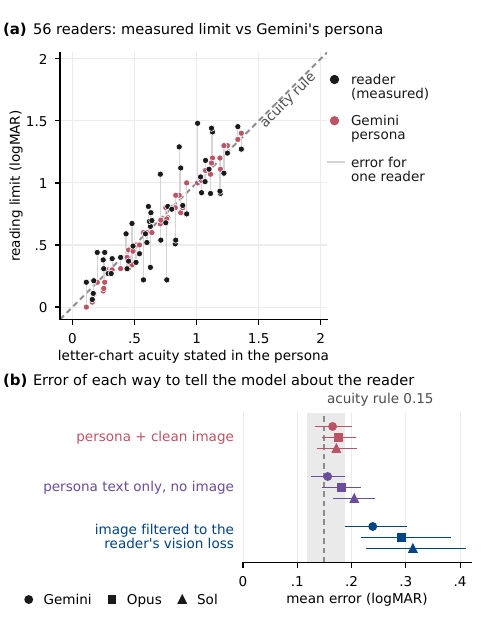}
\vspace{-1.6em}
\caption{\textbf{A reading persona copies the stated letter-chart acuity, so it misses readers whose reading limit is different.} (a)~Stated letter-chart acuity against reading limit, per reader (logMAR). Black: measured limit. Red: \gem with the persona. Dashed: acuity rule. (b)~Mean error of the frontier models for three ways of telling the model about the reader; none beats the acuity rule (dashed, 95\% interval).}
\label{fig:reading}
\vspace{-0.8em}
\end{figure}

\paragraph{Low-vision readers: two different acuities.} Xiong \etal~\cite{xiong2021lowvision} published vision and reading data for 58 people with low vision. We use the 56 whose vision lies inside the range where their filter method is valid. Two numbers describe each reader. The \emph{letter-chart acuity} is the smallest letter size that the reader can identify on an eye chart. The \emph{reading limit} is the smallest print size at which the reader can read sentences. Both are in logMAR units, where a larger value means larger print and 0.1 is one line on a chart~\cite{legge2007reading}. The two numbers often differ. For example, a reader with macular disease (damage to the centre of the retina) can identify single letters but fail to read sentences of the same size.

\paragraph{The task.} The readers took the MNREAD reading test in the original study; we use only their published results. The model takes the same kind of test, once for each reader's persona. It copies 24 sentences, one per print size from 2.0 to $-0.3$\,logMAR. We wrote these sentences ourselves, because the MNREAD sentences are copyrighted. Every image shows its sentence at the same legible pixel size, so a failure can only come from the persona, not from what the model can see; the image filter below tests the seeing route. The print size reaches the model only as a number in the prompt, or through an image filter. We score the reading limit as on the clinical MNREAD chart (supplement). The persona states the reader's letter-chart acuity, contrast sensitivity and diagnosis. A model that only copies the stated letter-chart acuity does as well as the acuity rule. Such a model fails for the readers whose two numbers differ.

\paragraph{The persona changes the answers, but toward the stated acuity.} Without a persona, the three frontier models copy every sentence correctly. With a persona, they stop at or near the stated letter-chart acuity: \gem stops within one chart line of it for 50 of 56 readers (\cref{fig:reading}a). Thus, the persona changes the answers, but the models only copy the acuity from the prompt. As a result, no persona predicts the reading limit better than the acuity rule, which is off by 0.15\,logMAR on average (\cref{fig:reading}b): \gem and \opus are slightly worse than the rule, and \sol is level with it (supplement).

\paragraph{The persona text holds the missing information, but the models do not use it.} Some readers read worse than their letter-chart acuity, and some read better. Does a persona show this? That is, when a reader reads worse than their acuity, does the model's answer for that reader's persona also read worse than the stated acuity? It does not. We correlate the two differences, the model's and the reader's, over the 56 readers (0 means no relation, 1 means full agreement). The correlation is at most 0.20 for every model, and its 95\% interval includes zero for every model, so we cannot tell it from no relation. Yet the persona text contains this information. The persona states three facts: letter-chart acuity, contrast sensitivity and diagnosis. A linear regression that uses only these three facts finds which readers read worse than their acuity: its correlation is 0.38\ci{0.19}{0.57}, clearly above zero. The regression cannot simply remember a reader, because we fit it on 55 readers, predict the remaining one, and repeat this for each of the 56 readers. Thus, the models mostly use only the stated acuity. For example, for readers with macular disease, all three personas read smaller print than the readers can. The open models mostly ignore the persona or give no answer (supplement).

\paragraph{A stronger prompt helps only by giving away the answer.} A worked example~\cite{natalie2025} shows one sentence at two print sizes that bracket the reader's limit. It lowers the error, but no more than a rule that clips the stated acuity to the two sizes. Step-by-step reasoning helps no model (supplement).

\paragraph{An image filter fixes the group average, not the individual.} A persona describes the vision loss in words; a filter puts it into the image. The filter starts from the reader's contrast threshold for each size of detail, which we compute from the reader's letter-chart acuity and contrast sensitivity~\cite{barten1999}. It removes the detail that the reader cannot see and adds noise at the threshold (supplement; similar in spirit to Wolff \etal~\cite{wolff2026lowpass}). On average, this filter works: the mean difference from the reader's limit is between $-0.01$ and $+0.11$\,logMAR in five of seven models. For individual readers, it does not. For each reader, we compare the model's reading limit on the filtered sentences with the reader's limit. In all seven models, this error is larger than the error of the acuity rule, by 0.09 to 0.19\,logMAR (one to two chart lines; \cref{fig:reading}b). The filter is built from only the acuity and contrast sensitivity that the persona also states. So it cannot show damage that these two numbers miss, such as the damaged centre of vision in macular disease, which makes sentences harder to read than single letters.

\begin{figure}[t]
\centering
\includegraphics[width=\columnwidth]{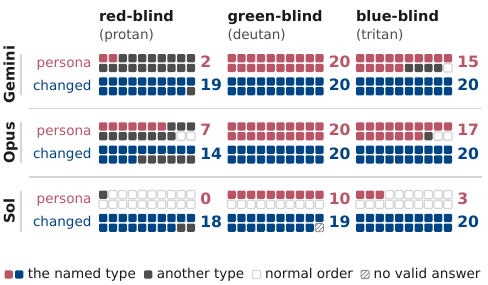}
\vspace{-1.6em}
\caption{\textbf{Colour-blind personas give a partial textbook answer that fails for red-blind viewers; changed colours work.} Each square is one of 20 cap layouts per type. Red: persona. Blue: no persona, colours changed as that type sees them. Filled: ordered like the named type; dark grey: like another type; open: like a normal viewer; hatched: no valid answer.}
\label{fig:d15}
\vspace{-0.8em}
\end{figure}

\paragraph{Colour-blind personas recite part of a textbook.} In the Farnsworth D-15 test, a viewer puts 15 coloured caps in order of hue. People with normal colour vision place them in a smooth circle. Colour-blind viewers confuse some hues, so their order crosses the circle, and the direction of the crossings shows the type: red-blind (protan), green-blind (deutan) or blue-blind (tritan). Published group scores of 39 measured colour-blind viewers (12 protans, 23 deutans and 4 tritans) define the pattern of each type~\cite{vingrys1988d15}. Individual orders are not published, so we compare the model with each type, not with individual viewers. The persona names the type and its missing cone. For each type, the model orders 20 random cap layouts. We score each order with the standard method of Vingrys and King-Smith~\cite{vingrys1988d15}, which gives the direction and the size of the crossings. A layout counts as correct when its crossings point in the direction of the named type, so the maximum score is 60 (20 per type).

Two simple controls set the bar. Without a persona, every frontier model orders the caps like a normal viewer, so it scores 0 of 60. The textbook answer for each stated type needs no model: it orders the caps along the one colour direction that this type still sees. It scores 60 of 60, and its confusion angles lie close to the measured group means. For comparison, a single deutan order given for every type scores 20 of 60.

With a persona (\cref{fig:d15}), \gem and \opus score 37 and 44 of 60, well below the textbook answer for the stated type, and unevenly across types. For deutans, both score 20 of 20. For tritans, they score 15 and 17. For protans, they score only 2 and 7, because they usually give the deutan order instead. \sol's personas mostly order the caps like a normal viewer, so \sol scores 13 of 60, below even the single deutan order. In contrast, when we change the colours as each type sees them~\cite{machado2009cvd,brettel1997cvd} and give no persona, the models score 54 to 59 of 60.

\begin{figure}[t]
\centering
\includegraphics[width=\columnwidth]{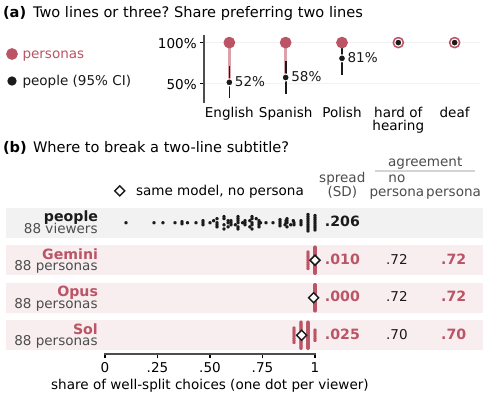}
\vspace{-1.6em}
\caption{\textbf{Subtitle personas answer like the model, not like the viewer.} (a)~Share of viewers who prefer two-line to three-line subtitles (95\% intervals; all 19 deaf and hard-of-hearing viewers chose two lines). (b)~A different task: for 30 two-line subtitles, viewers chose between a well-split and a poor line break. Each dot is one person (black) or that person's persona (red). Diamonds: model without a persona. Right: spread across viewers (SD), and agreement, the share of frames on which the model chose the same break as the viewer, with no persona and with that viewer's persona.}
\label{fig:groups}
\vspace{-0.8em}
\end{figure}

\paragraph{Subtitle viewers test hearing and language, not sight.} The SURE study~\cite{szarkowska2018suredata,szarkowska2018fast,gerbermoron2018linebreaks,gerbermoron2018segmentation,szarkowska2019lines} recorded 88 to 94 viewers per task in five groups: hearing native speakers of English, Polish and Spanish, and viewers who are deaf or hard of hearing. The viewers did four tasks. In the first, they chose the better of two line breaks for 30 subtitled TV frames; the \emph{well-split line} keeps a phrase together on one line. In the other tasks, they rated film clips whose subtitles had good or bad line breaks, two or three lines, and slow or fast speed. They rated, for example, speech--subtitle mismatch and annoyance, and stated which subtitles they preferred. We make one persona for each participant. Each persona states these facts in a fixed template (supplement). Each persona gets the same task as its participant. \gem receives the video with sound; \opus and \sol receive frames that show every subtitle.

\paragraph{The persona does not bring the model closer to the person.} The line-break task uses still screens, which we rebuilt from the experiment file with a substitute font and both left--right orders (supplement). In this task, people choose the well-split line in 72\% of trials, and they disagree with each other. For each participant, we compare their own 30 choices with the choices of their persona and with the choices of the same model without a persona. The agreement is the fraction of the 30 frames on which the two choose the same line break. The persona does not raise this agreement in any frontier model (\cref{fig:groups}b; changes $-0.001$ to $+0.002$, all intervals include zero). The personas agree with people only about as often as the rule ``always choose the well-split line.'' Across items, the choices of the \gem and \sol personas correlate with people's at only .05 and .14, while two halves of the human sample correlate at .84; the \opus personas choose the well-split line in every trial. The personas also erase the differences between individuals. People differ: some choose the well-split line in 23\% of trials, others in all of them. The personas barely differ: the spread (standard deviation) of their shares across participants is 8 to 20 times smaller than people's, and zero for \opus (\cref{fig:groups}b). This holds when we sample five answers per item or ask for choice probabilities (supplement). People differ in a stable way (split-half reliability .88), but the stated facts do not predict who chooses the well-split line more often (supplement), so here the narrow spread is a mismatch with people, not evidence that the personas ignore useful facts. Real examples do carry this information. When the model also sees 10 of the person's own choices and predicts the other 20, it agrees with the person about as often as a simple predictor that uses the same 10 choices (.72--.76 against .74), and \gem and \opus recover the human spread. Two controls show which part helps: without the examples, the same persona agrees with people less often, and without the persona, the examples alone do as well. Thus, the examples carry the information, not the persona (supplement).

\begin{figure}[t]
\centering
\includegraphics[width=\columnwidth]{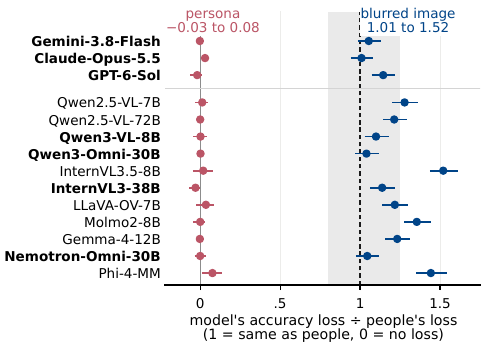}
\vspace{-1.6em}
\caption{\textbf{A text description of blur has no effect in any of 14 models; blurred photos lower accuracy in all of them.} Model's accuracy loss divided by people's (1 = same; 0 = none; 95\% intervals). Red: clean photo with the persona. Blue: blurred photo. Grey band: 0.80--1.25. Bold: as large as people's after Holm correction.}
\label{fig:blur}
\vspace{-1.2em}
\end{figure}

\paragraph{The differences between groups disappear.} A group difference is the difference between two groups on one measure. Deaf and hard-of-hearing viewers prefer two-line subtitles far more often than English viewers: all 19 against 14 of 27 (52\%), a difference of 48 percentage points. Every persona, and every frontier model without a persona, prefers two lines, so the difference is zero (\cref{fig:groups}a). The model's default thus matches the deaf and hard-of-hearing viewers and misses the English viewers, about half of whom accept three lines. We test 33 group comparisons from the study. A comparison is fair only if the model notices what changes in it without a persona. No frontier model notices a change in subtitle speed, and some miss other changes, so we drop those comparisons. This leaves 22 of the 33 comparisons for each model. In 7 of these 22, the human difference is clear (its 95\% interval excludes zero). For example, bad line breaks make subtitles harder for Spanish viewers than for English viewers. No persona reproduces any of these 7 differences. In most of them, the personas show almost no difference between the groups, and in a few, the difference points the other way (supplement). Over all 33 comparisons, the persona differences do not follow the human ones: their correlation is between $-0.17$ and $+0.23$, and every interval includes zero.

\takeaway{Personas of measured people make the model use the stated facts, a stereotype, or its own default. In reading and subtitles, no persona predicts people better than its task-specific control, and no persona keeps the differences between subtitle-viewer groups.}

\begin{figure}[t]
\centering
\includegraphics[width=\columnwidth]{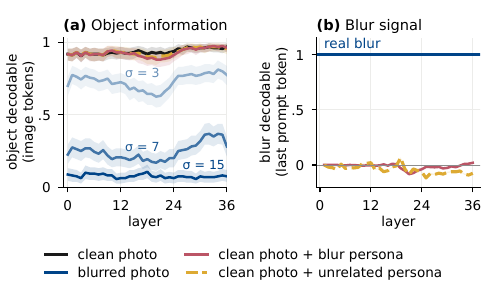}
\vspace{-1.6em}
\caption{\textbf{Inside Qwen3-VL-8B, the blur persona leaves the image information intact.} Linear probes of the hidden states on held-out photos, per layer (bands: 95\% intervals). (a)~The object is as decodable from the image tokens with the blur persona (red) as for the clean photo (black) or an unrelated persona of the same length (yellow, dashed); blurred photos (blue) remove more of it as the blur grows. (b)~A probe that tracks real blur (= 1) at the last prompt token reads almost none of the stated blur, no more than for the unrelated persona.}
\label{fig:internal}
\vspace{-1.2em}
\end{figure}

\begin{figure*}[t]
\centering
\includegraphics[width=\textwidth]{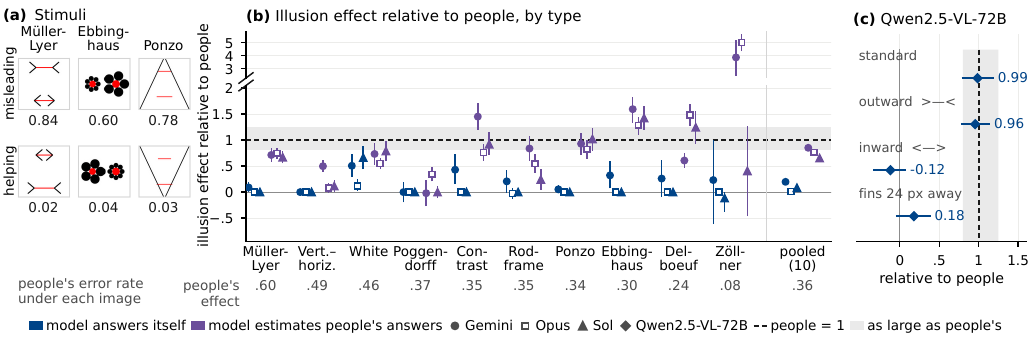}
\vspace{-1.4em}
\caption{\textbf{Models know how often illusions fool people, but their own answers are rarely fooled.} (a)~Example images; numbers are human error rates. (b)~Illusion effect, relative to people, when \gem, \opus and \sol answer themselves (blue) and estimate people's answers (purple). Grey band: 0.80--1.25. Bottom row: human effect. (c)~Qwen2.5-VL-72B matches people on M\"uller-Lyer (0.99), but only outward fins mislead it, while both fin types mislead people.}
\label{fig:illusion}
\vspace{-1.3em}
\end{figure*}

\subsection{Can a description replace a changed image?}\label{sec:proxy}
Why do personas fail? A persona of a person with a visual limit must tell the model about the limit and also make the model lose the information that the limit removes. To test the second job, we use a task where we can also remove the information from the image. The human reference is sighted people who view the changed images.

\paragraph{The blur test.} In the model-vs-human benchmark~\cite{geirhos2021}, four observers with normal vision named the object in greyscale photos from 16 classes. Each photo appeared for 200\,ms at one of eight blur levels. We give each model 160 photos per level in two ways. Either the model sees the blurred photo, or it sees the clean photo with a persona that states the acuity that matches the blur. We convert blur to acuity by assuming that the photo covers 3$^\circ$ of the visual field. The acuity is in Snellen notation: 20/165 means that the person sees at 20 feet what a typical viewer sees at 165 feet (0.92\,logMAR). \Cref{fig:teaser} (top right) shows an example.

\paragraph{Describing blur has no effect.} The blur persona has no effect on recognition in any of the 14 models (\cref{fig:blur}). At the strongest stated loss, acuity 20/940, people are at chance, but the models with the persona still name 84 to 98\% of the photos correctly (\cref{fig:teaser}, bottom). The models say that they are less confident, but their answers do not change. For \gem, the result is the same with the two persona prompts of Natalie \etal~\cite{natalie2025} (a minimal one and one that states the diagnosis), with eight example human answers, and with step-by-step reasoning. Personas that describe noise or four other image changes have no effect either (supplement). The models know the effect that they do not show. Asked instead to estimate how accurately people with each stated acuity name the objects, the three frontier models give a loss of 0.80 (\gem), 0.98 (\opus; as large as people's) and 0.82 (\sol) of the human loss.

\paragraph{Blurring the photo lowers recognition.} Blurred photos lower accuracy in every model, and in seven models, including the three core frontier models, the loss is as large as people's (\cref{fig:blur}). The blur at which accuracy falls halfway to chance gives the same picture: it matches people's in four models, including \gem and \opus, and no model tolerates more blur than people (supplement). Models and people do not always fail on the same photos, but neither do two people: the agreement between a model and people on which photos are missed is at most $\kappa=.35$, and between two people it is $\kappa=.40$ ($\kappa=0$: chance agreement; 1: full agreement).

\paragraph{Inside the model, the description changes nothing that blur changes.} To see why the description fails, we look inside open models that do the blur task without a persona. First, the answer probabilities. In four models, the probability of the correct object is .94--.95 with the blur persona, the same as without it; when the photo itself is strongly blurred ($\sigma=15$ pixels), it falls to .05--.10, near chance. Second, the hidden states, that is, the output of each transformer block, at every layer of three models. We place the persona before the image, so that the image tokens can take it into account. We train linear probes on the hidden states of some photos and test them on others. The object can be read from the hidden states with the persona as well as without it, while blur removes most of it (\cref{fig:internal}a). A second probe, trained on the hidden states of blurred photos, reads how blurred the photo is. Its reading rises one-to-one with real blur, but with the stated blur of the persona it rises by at most 0.17 of that (Gemma-4-12B), 0.08 (Qwen3-VL-8B) and 0.04 (InternVL3.5-8B) (\cref{fig:internal}b). The persona does change the hidden states, but almost perpendicular to the change that blur makes (cosine 0.1--0.2). To check that the probes can see a text effect, we add an instruction that does change the answers (``name a different object''): it changes 82--99\% of the answers, and the probes detect the change. Thus, in these models the persona changes neither the image information nor the decision; it adds only a small drop in confidence. \Cref{fig:internal} shows Qwen3-VL-8B; Gemma-4-12B and InternVL3.5-8B give the same result (supplement).

\takeaway{The information that a visual limit removes reaches the model through the image, not through a description. A description of the limit leaves the answers unchanged.}

\subsection{What does text give the model instead?}\label{sec:stated}
If a persona does not change what the model sees, what does text change? A diagnostic test answers this: the model estimates how a group of people answered.

\paragraph{Models know how often illusions fool people, but are rarely fooled.} Makowski \etal~\cite{makowski2023illusion} recorded fast choices from 256 people on ten types of visual illusion, such as ``Which red line is longer?'' We remake the images with the same software~\cite{makowski2021pyllusion}. The \emph{illusion effect} is the error rate when the context misleads minus the error rate when the context helps. In their own answers, \opus shows 1\% of people's illusion effect, \sol 8\% and \gem 20\% (\cref{fig:illusion}b). We also ask each model, in the same reply, for the percentage of adults who choose each option. These estimates show 65\% (\sol), 76\% (\opus) and 85\% (\gem) of people's illusion effect. They also find the images that mislead people better than the models' own answers do (correlation .66--.76 against .10--.30).

\paragraph{Estimates do not replace a simulated viewer.} Three limits keep estimates from standing in for answers. First, an estimate is a single number for a group, such as ``adults''; it gives no individual answers and no spread between people, which a user study needs. Second, its errors are large and cannot be found without human data: \gem and \sol estimate almost no Poggendorff effect, although it misleads people strongly, and \opus and \gem estimate the weak Z\"ollner effect four to five times too large. Third, estimates are weaker for single items than for whole curves: for blur, per-photo estimates recover 0.42 to 0.74 of the human loss, against 0.80 to 0.98 for one curve over all levels (supplement). An estimate shows what the model knows about people; it does not replace measuring them.

\paragraph{A matching average can hide a different cause.} Of 11 open models, only Qwen2.5-VL-72B passes our control images, and its M\"uller-Lyer effect matches people's (relative effect 0.99). Control figures show why (\cref{fig:illusion}c). In people, both outward and inward fins mislead; in the model, only outward fins do ($-0.12$ for inward fins). Asked for lengths in pixels, it adds about half a pixel per pixel of outward fin, and moving the fins 24 pixels away from the line removes 82\% of its effect. The model counts part of the fins as line, which happens to match people.

\paragraph{A persona can turn an illusion on.} A young-adult persona raises \gem's illusion effect from 20\% to 84\% of the human value, and its errors fall on the images that mislead people (correlation .52 across images, against .66 for its estimates); the personas of \opus and \sol barely change their answers. Text can thus produce a human-like error when the model knows which way people err; this is recall of the known effect, not perception of the image. An age persona does not reproduce the measured age differences (supplement).

\takeaway{Text lets a model recall how people answer, but not see what they see: in its own answers, the model is rarely fooled by the illusions that fool people.}

\section{Discussion}\label{sec:discussion}
\paragraph{Telling is not seeing.} A persona can make a model say what a person would say, but only when the model already knows that answer: the direction of an illusion, the textbook deutan order, or the rule that the reading limit equals the stated acuity. It never makes the model see what the person sees. When the answer depends on what the person's vision lacks, or on differences for which the model has no rule, the persona changes nothing. Inside open models, the description changes neither the image information nor the decision. A simulator need not change perception: a decision rule built from the model's own estimates could reproduce group-level errors. Current personas do neither. The frontier models' answers show the same pattern. The tests also differ in what the stated facts allow: in reading, they predict which readers depart from the rule, which the personas ignore; in the line-break task, they predict no individual choices; in colour sorting, the textbook order is a ceiling.

\paragraph{Limitations.}\label{sec:limits} We test short template personas; richer descriptions are a natural next test. The blur test is a controlled proxy, not clinical low vision, and people saw each photo for 200\,ms. Several reference groups are small (4 blur observers, 4 tritans, 9 to 27 viewers per subtitle group). Our inputs differ from what people saw: the reading sentences are always legible, the line-break screens use a substitute font, and \opus and \sol see frames without sound.

\paragraph{How to evaluate a simulated viewer.} We suggest four checks: compare the persona with people who have the condition; compare it with a rule from the stated facts; check items and people, not only averages; and keep the model's estimates apart from its own answers. A persona can point to a possible barrier, but only tests with people who have that need can show whether the barrier is real.

{
    \small
    \bibliographystyle{ieeenat_fullname}
    \bibliography{refs}
}

\end{document}